\documentclass[sigconf]{acmart}

\def\BibTeX{{\rm B\kern-.05em{\sc i\kern-.025em b}\kern-.08emT\kern-.1667em\lower.7ex\hbox{E}\kern-.125emX}}
    
\usepackage{booktabs}
\usepackage{amsthm}
\usepackage{algorithm}
\usepackage{algorithmic}

\begin{document}

%
\title{CPM-sensitive AUC for CTR prediction}

%
\author{Zhaocheng Liu}
\email{zhaocheng.liu@mobvista.com}
\affiliation{%
  \institution{Mobvista. Inc}
  \city{Chaoyang}
  \state{Beijing}
  \country{China}}
  
\author{Guangxue Yin}
\email{guangxue.yin@mobvista.com}
\affiliation{%
  \institution{Mobvista. Inc}
  \city{Chaoyang}
  \state{Beijing}
  \country{China}}
%

%
\begin{abstract}
  The prediction of click-through rate (CTR) is crucial for industrial applications, such as online advertising. 
  AUC is a commonly used evaluation indicator for CTR models. 
  For advertising platforms, online performance is generally evaluated by CPM. 
  However, in practice, AUC often improves in offline evaluation, but online CPM does not. 
  As a result, a huge waste of precious online traffic and human costs has been caused. 
  This is because there is a gap between offline AUC and online CPM. 
  AUC can only reflect the order on CTR, but it does not reflect the order of CTR*Bid. 
  Moreover, the bids of different advertisements are different, so the loss of income caused by different reverse-order pair is also different. 
  For this reason, we propose the CPM-sensitive AUC (csAUC) to solve all these problems. 
  We also give the csAUC calculation method based on dynamic programming. 
  It can fully support the calculation of csAUC on large-scale data in real-world applications.


\end{abstract}

%
%

%

%

%
\maketitle

\section{Introduction}
The prediction of click-through rate (CTR) is crucial for online advertising. 
For each request, we need to choose the ads that maximize our benefits from a bunch of candidate ads to return. 
The bid and CTR of each ad varies on different traffic. 
Therefore, in order to maximize the expected revenue, the common practice is to use CTR*bid as the basis for rank. 
We need to constantly optimize the CTR model to make the online rank more accurate.

To determine whether the new iterated model is better than the baseline, the usual approach is to evaluate the model offline first. 
Compare the AUC of the old and new models on the validation data set. 
If the auc is improved, the online A/B test experiment can be carried out to verify the online cost per mille (CPM) improvement.

However, it often happens that offline auc has improvement, while online CPM has not. 
This is due to the gap between offline auc evaluation and online CPM performance. 
Traditional AUC is the probability of correct ranking of a random “positive”-“negative” pair. 
That is to say, only the order of CTR is considered, while the order of CTR*bid is not. 
Moreover, different reverse-order pairs will cause different income losses. 
AUC also cannot reflect this. 

A lot of work has been done in the field of CTR estimation, but most of them focus on the model structure (FM\cite{rendle2010factorization}, Wide\&Deep\cite{cheng2016wide}, PNN\cite{qu2016product}, DeepFM\cite{guo2017deepfm}, DIN\cite{Zhou2017Deep}, DIEN\cite{Zhou2018Deep} etc). 
There are also some efforts to solve real-world common problems (ESSM\cite{Ma2018Entire}, DeepMatch\cite{zhu2018learning}, Rocket Training\cite{zhou2017Rocket} etc).
However, none of the above work attempts to solve the fundamental problem——gap between evaluation indicator and online performance. 
The GAUC\cite{zhu2017optimized} only guarantees the consistency of off-line and on-line in sample distribution to a certain extent. 
But GAUC can't eliminate the gap mentioned above. 

In order to solve the above problem, we propose CPM-sensitive AUC (csAUC). 
In the next section you will see its definition and analysis. 
In order to use csAUC in real-world applications, we give a detailed calculation of csAUC in the third section.

\section{Our Approach}
\subsection{Definition of csAUC}
In dataset D, a positive sample $x_{pos}$ and a negative sample $x_{neg}$ are randomly selected. The estimated values of the two samples in the model are $pCTR_{pos}$ and $pCTR_{neg}$. The value of AUC indicates the probability that $pCTR_{pos}$ is greater than $pCTR_{neg}$. 
The Definition of AUC is as follows:
\begin{definition}\label{traditional-auc-def}Using $pCTR$ as rank basis, AUC is the probability of correct ranking of a random $(x_{pos}, x_{neg})$ pair.\end{definition}

In our csAUC: 
\begin{itemize}
  \item \textbf{Samples are multilevel}. The level of negative sample is the lowest, and the level of positive sample is determined according to its corresponding bid.  The higher the bid is, the higher the level of positive sample is. 
  \item \textbf{The revenue of every negative sample is 0}. The revenue of a positive sample equals its bid.
\end{itemize}

Now a high-level sample $x_h$ and a low-level sample $x_l$ are randomly selected from dataset D. Define our revenue (Rev) of $(x_h, x_l)$ is as follows:
$$Rev(x_{h}, x_{l}) = \begin{cases}bid_{h} & pCTR_h*bid_h >= pCTR_l*bid_l\\T(x_l) & otherwise\end{cases}$$
$$T(x_i)=\begin{cases}0 & if\ x_i\ is\ a\ negative\ sample\\bid_i &otherwise\end{cases}$$

Using Rev function define csAUC as [DEF \ref{csauc-def}]
\begin{definition}\label{csauc-def}CsAuc of dataset D is $\frac{\sum_{(x_h, x_l)\in D} Rev(x_h, x_l)}{\sum_{(x_h, x_l)\in D} bid_h}$.\end{definition}

\subsection{Anlysis of csAUC}
When the pCTR of any positive sample is greater than the pCTR of every negative sample, the AUC is equal to 1. 
At this time, the csAUC is not necessarily equal to 1. 
\begin{corollary}\label{csAUCequals1}csAUC is equal to 1 \textbf{if and only if} the pCTR of any positive sample is greater than the pCTR of all negative samples, and positive samples are internally reversed by bid. \end{corollary} 

Traditional AUC can only measure the discrimination of positive and negative samples, it is far from enough. 
For example, our validation set is shown in [TAB \ref{tab:data-demo}]. 
Suppose we have six different CTR models that need to be validated on this validation dataset. 
Using $pCTR*bid$ as rank basis, the results of simulated rank are Seq[1-6] which are shown in [TAB \ref{tab:validation-demo}]. 
The AUC and csAUC for these squences are shown in [TAB \ref{tab:demo-auc-csauc}].  

Seq[1-4] successfully separate positive and negative samples, but the bias of ordering inside positive samples causes the values of csAUC to be different. Specifically, seq2 is the most profitable, seq1 loses a lot of revenue, and the benefits of seq3 and seq4 are equal. This reflects that the value of csAUC is closely aligned with offline cpm performance, while AUC is not. Moreover, Seq[5,6] shows that AUC is sometimes negatively correlated with offline cpm performance. 
Since in the offline data, the label of a sample is equal to 1 or 0, thus csAUC defined in [DEF \ref{csauc-def}] represents offline CPM performance. Therefore, csAUC is a better offline evaluation indicator than AUC.
\begin{table}
  \caption{Data demo}
  \label{tab:data-demo}
  \begin{tabular}{ccc}
    \toprule
    SampleID&Bid&Label\\
    \midrule
    A & 100 & 1\\
    \cline{1-3}
    B & 4 & 1\\
    \cline{1-3}
    C & 3 & 1\\
    \cline{1-3}
    D & 2 & 1\\
    \cline{1-3}
    E & 999 & 0\\
  \bottomrule
\end{tabular}
\end{table}

\begin{table}
  \caption{Sequences demo}
  \label{tab:validation-demo}
  \begin{tabular}{c|c|c|c|c|c}
    \toprule
    Seq1&Seq2&Seq3&Seq4&Seq5&Seq6\\
    \midrule
    D & A & A & A & B & A\\
    C & B & C & B & C & B\\
    B & C & B & D & D & E\\
    A & D & D & C & E & C\\
    E & E & E & E & A & D\\
  \bottomrule
\end{tabular}
\end{table}

\begin{table}
  \caption{AUC and csAUC for Seqs in Table 1}\label{tab:demo-auc-csauc}
  \begin{tabular}{ccccccc}
    \toprule
    &Seq1&Seq2&Seq3&Seq4&Seq5&Seq6\\
    \midrule
    AUC & 1 & 1 & 1 & 1 & 0.9 & 0.8\\
    csAUC & 0.2976 & 1 & 0.9976 & 0.9976 & 0.069 & 0.988\\
  \bottomrule
\end{tabular}
\end{table}

\subsection{gcsAUC}
In order to ensure that the sample distribution is more consistent between offline and online evaluation, we can do group like GAUC\cite{zhu2017optimized} first, and then calculate csAUC in group.
We call it gcsAUC. 
gcsAUC is the maximum simulation of online CPM in offline evaluation. 

\subsection{Overall Offline Evaluation Metrics}
We generally use AUC, COPC (click-over-predicted-click), gcsAUC, ROPR (revenue-over-predicted-revenue) for offline evaluation. 
\begin{itemize}
  \item AUC is used to evaluate the order of CTR.
  \item COPC is used to assess whether the overall estimate is high or low.
  \item gcsAUC reflects the performance of offline simulated CPM. 
  \item ROPR reflects whether the expected income estimate is high or low. 
\end{itemize} 
The formula for calculating COPC and ROPR of dataset D is as follows: 
$$COPC(D) = \frac{\sum_i y_i}{\sum_i pCTR_i}$$
$$ROPR(D) = \frac{\sum_i y_i * bid_i}{\sum_i pCTR_i * bid_i}$$

\section{Implementation of csAUC}
Since the samples are multilevel, the computational complexity of csAUC is much larger than that of AUC. 
We take a two-level bucket to record the required statistics. 
Specifically: 
\begin{itemize}
  \item The first-level bucket represents a combination of label and bid. All negative samples are placed under barrel No. 0 of the first level. For positive samples, bid is divided into buckets. The bigger the bid is, the bigger the bucket id. 
  \item The second bucket represents the bucket of pCPM, calculates the pCPM of all samples in the validation set, and normalizes it by Min-Max norm. Multiply the value of norm by 1E5 to get the second-level bucket number.
\end{itemize}
The Min-Max norm is as follows:
$$pCPM_i' = \frac{pCPM_i - min\{pCPM_1, .. , pCPM_n\}}{max\{pCPM_1, .., pCPM_n\} - min\{pCPM_1, .., pCPM_n\}}$$

After calculating the bucket number of each sample, the COUNT operation is performed inside the bucket. 
So far we have triple data like (level\_1, level\_2, cnt\_num). 

Suppose there are $l_1$ buckets in the first-level and $l_2$ buckets in the second-level. 
The time complexity of the most direct method for calculating csAUC from triple data is $O(C_{l_1}^2*l_2)$. 
Based on dynamic programming, we reduce the time complexity to $O(l_1 * l_2)$. See [ALG \ref{alg:Cal-csAUC}]. 

\begin{algorithm}[htb]
\caption{ Calculation of csAUC.} 
\label{alg:Cal-csAUC} 
\begin{algorithmic}[1] 
\REQUIRE ~~\\ 
The set of predict sample $(value, ecpm, cnt)$, $P_n$;\\
The max level of value $ml_{v}$ and max level of ecpm $ml_c$
\ENSURE ~~\\ 
cpm-sensitive AUC, $csAUC$; 
\STATE Init $grid, left, middle, right$ as zero $ml_v \times ml_c $ matrix, $ls$ as zero $ml_v \times 1$ matrix
\label{ code:fram:init }
\STATE feed $P_n$ into $gird$ as
\FOR{$(value, ecpm, cnt) \in P_n$}
\STATE $grid_{value, ecpm} += cnt$
\STATE $ls_{value} += cnt$
\ENDFOR
\STATE construct auxiliary variables
\FOR{$v \in range(ml_v)$}
\STATE $tmp = 0$
\FOR{$c \in range(ml_c)$}
\STATE $tmp += grid_{v, c}$
\STATE $left_{v, c} = (v == 0 ? 0 : left_{v-1, c}) + tmp$
\STATE $middle_{v, c} = (v == 0 ? 0 : middle_{v-1, c}) + grid_{v, c}$
\ENDFOR
\ENDFOR
\FOR{$v \in range(ml_v)$}
\STATE $tmp = 0$
\FOR{$c \in reverse(range(ml_c))$}
\STATE $right_{v, c} = (v == 0 ? 0 : right_{v-1, c}) + tmp$
\ENDFOR
\ENDFOR
\STATE calculate reward
\STATE $tmp = 0, reward_{max} = 0.0$
\FOR{$v \in range(ml_v)$}
\STATE $reward_{max} += ls_v * tmp * v$
\STATE $tmp += ls_v$
\ENDFOR
\STATE $reward_{rank} = 0.0$
\FOR{$v \in range(1,ml_v)$}
\FOR{$c \in range(ml_c)$}
\STATE $reward_{rank} += (c == 0 ? 0 : left_{v-1, c-1}) * grid_{v, c} * v$
\STATE $reward_{rank} += ((right_{v-1, j} - (c == ml_c - 1 ? 0 : right_{i-1, j+1})) + middle_{i-1, j})*grid_{v, c} * v * 0.5$
\STATE $reward_{rank} += (c == ml_c - 1 ? 0 : right_{v-1, c+1}) * grid_{v, c}$
\ENDFOR
\ENDFOR
\RETURN $reward_{rank} / reward_{max}$; 
\end{algorithmic} 
\end{algorithm}

\section{CONCLUSION}
We propose the definition, analysis and fast calculation method of csAUC. Compared with AUC, csAUC relies on CTR*Bid as the sorting criterion, and is more sensitive to the loss of income caused by different reverse-order pairs. 
At the same time csAUC can be combined with gauc, we call it gcsAUC. 
gcsAUC is basically the maximum simulation of online CPM in offline evaluation. 

In addition, we also present the whole set of offline indicators that we are using. 
Each indicator provides its own insights for understanding the effects of our models. 
%

%
\bibliographystyle{ACM-Reference-Format}
\bibliography{refs}

\end{document}